\pdfoutput=1
\documentclass[11pt]{article}

\usepackage{acl}
\usepackage{float}
\usepackage{times}
\usepackage{graphicx}
\graphicspath{ {./images/} }
\usepackage{array}
\usepackage{latexsym}
\usepackage{amsfonts}
\usepackage{booktabs}
\usepackage{multirow}
\usepackage{pifont}
\usepackage{bbding}
\usepackage{nameref}
\usepackage{caption}
\usepackage{subcaption}
\usepackage{siunitx}
\usepackage{amsmath}

\usepackage{xcolor}
\usepackage{amsmath}
\usepackage{hyperref}
\usepackage{xcolor}
\definecolor{darkblue}{rgb}{0, 0, 0.5}
\hypersetup{colorlinks=true, citecolor=darkblue, linkcolor=darkblue, urlcolor=darkblue}
\definecolor{IncorrectText}{HTML}{ff0000}
\definecolor{IncorrectPlace}{HTML}{3366ff}
\newcommand{\incorrettext}[1]{\textcolor{IncorrectText}{#1}}
\newcommand{\incorretplace}[1]{\textcolor{IncorrectPlace}{#1}}
\newcommand{\circleone}{\textrm{\ding{172}}}
\newcommand{\circletwo}{\textrm{\ding{173}}}
\newcommand{\circlethree}{\textrm{\ding{174}}}

\newcommand{\codesnip}[1]{\small\texttt{#1}\normalsize}
\newcommand{\numwithstd}[2]{#1\small$\pm#2$\normalsize}
\newcommand*{\tableind}{\hspace*{0.5cm}}%
\newcommand{\thincross}{\textrm{\ding{61}}}
\newcommand{\checkding}{\textrm{\ding{51}}}
\newcommand{\codetable}[1]{\small\texttt{#1}\normalsize}
\newcolumntype{P}[1]{>{\raggedright\arraybackslash}p{#1}}
\usepackage[T1]{fontenc}
\usepackage[utf8]{inputenc}

\usepackage{microtype}

\title{Reading StackOverflow Encourages Cheating: Adding Question Text Improves Extractive Code Generation}

\author{Gabriel Orlanski \\
  Rensselaer Polytechnic Institute \\
  \texttt{orlang2@rpi.edu} \\\And
  Alex Gittens \\
  Rensselaer Polytechnic Institute\\
  \texttt{gittea@rpi.edu} \\}

\begin{document}
\maketitle
\begin{abstract}
Answering a programming question using only its title is difficult as salient contextual information is omitted. Based on this observation, we present a corpus of over 40,000 StackOverflow question texts to be used in conjunction with their corresponding intents from the CoNaLa dataset~\citep{yin2018learning}. Using both the intent and question body, we use BART to establish a baseline BLEU score of 34.35 for this new task. We find further improvements of 2.8\% by combining the mined CoNaLa data with the labeled data to achieve a 35.32 BLEU score. We evaluate prior state-of-the-art CoNaLa models with this additional data and find that our proposed method of using the body and mined data beats the BLEU score of the prior state-of-the-art by 71.96\%. Finally, we perform ablations to demonstrate that BART is an unsupervised multimodal learner and examine its extractive behavior.\footnote{https://github.com/gabeorlanski/stackoverflow-encourages-cheating}
\end{abstract}

\section{Introduction}
\indent The goal of semantic parsing is to translate a Natural Language(NL) utterance to its logical components. There is a large body of research on applying semantic parsing for source code generation in a multitude of domain specific languages such as lambda calculus and SQL \citep{dahl-etal-1994-expanding,zelleMooneyParseDB,zettlemeyorLearningtoMap,ling-etal-2016-latent,xiao-etal-2016-sequence,rabinovich-etal-2017-abstract,dong-lapata-2018-coarse,guo-etal-2019-towards,hwang2019comprehensive,tabassum-etal-2020-code}. However, the task of translating an NL utterance to a general-purpose programming language has proven to be more challenging. A significant issue contributing to this is the difficulty in acquiring quality data due to the necessary domain knowledge needed in the annotation process. 

\begin{figure}[ht]
    \centering
    \includegraphics[width=\linewidth,keepaspectratio]{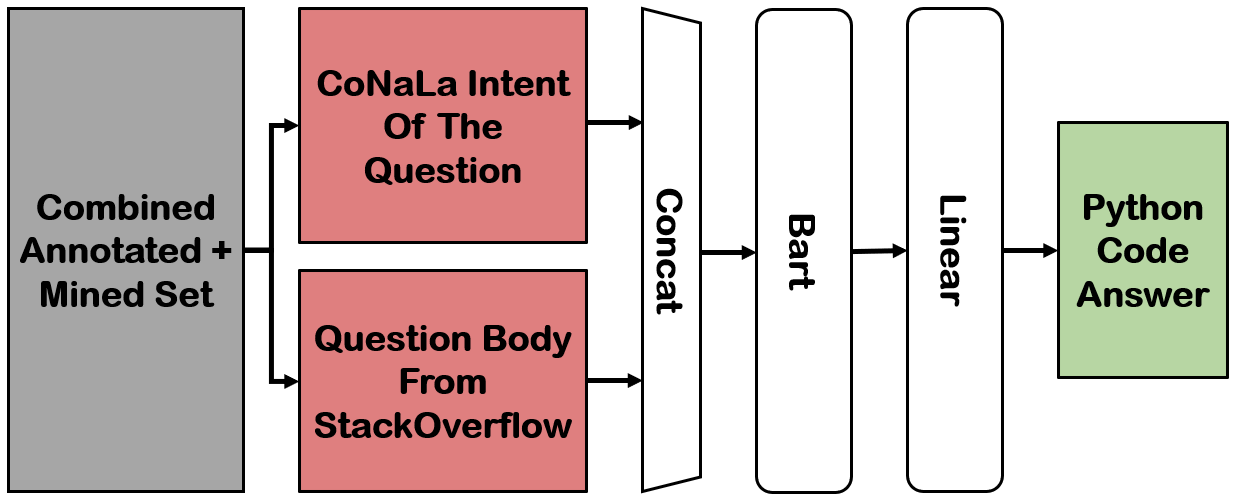}
    \caption{Overview of our approach. From the combined annotated + mined set, we concatenate the intent and question body for inputs to BART\citep{lewis-etal-2020-bart} and use beam search for generation.}
    \label{fig:approach}
\end{figure}
\indent Despite this, the past few years have seen a large number of datasets released for different text-to-code related tasks \citep{ling-etal-2016-latent,iyer-etal-2018-mapping, 10.1145/3178876.3186081,yu-etal-2018-spider, DBLP:journals/corr/abs-2102-04664}. Some datasets such as CodeSearchNet \citep{husain2019codesearchnet} contain snippets from a multitude of different languages. Others focus on distinct tasks within a specific language, such as JuICe \citep{agashe-etal-2019-juice}, which contains executable Python programming assignments. Utilizing these corpora, prior works \citep{suhr-etal-2018-learning, yin-neubig-2017-syntactic,yin-neubig-2018-tranx,Sun_Zhu_Xiong_Sun_Mou_Zhang_2020,hayati-etal-2018-retrieval,yin-neubig-2019-reranking, xu-etal-2020-incorporating,Drain2021GeneratingCW} have found success with a large variety of model architectures. These methods, however, struggle with domain agnostic open-ended code generation in general-purpose languages. One idea to combat this is to utilize large pretrained language models. 

\indent Transformers \citep{vaswani2017attention} have demonstrated that they can both be few-shot \citep{brown2020language} and unsupervised multitask \citep{radford2019language} learners. They have been successfully applied to programming language tasks. CodeBERT achieved strong performance on the CodeSearchNet task through pretraining on bimodal NL comment and code pairs\citep{feng-etal-2020-codebert}, while \citet{Sun_Zhu_Xiong_Sun_Mou_Zhang_2020} used abstract syntax trees(AST) and transformers to achieve state of the art performance on the HearthStone benchmark\citep{ling-etal-2016-latent}. \citet{roziere2021dobf} proposed the deobfuscation pretraining task to incorporate structural features of code into transformer models without the use of ASTs. More recently, \citet{shin2021constrained} explored the capabilities of large pretrained language models to be few-shot semantic parsers. 

\indent Yet open-domain programming question answering on sites such as StackOverflow(SO)\footnote{https://stackoverflow.com/} has remained an elusive goal. \citet{yin2018learning} created an annotated dataset with the site in which the intent and answer snippet pairs were automatically mined from the question. They then had crowd workers rewrite the intents to reflect the corresponding code better. Currently, state-of-the-art was achieved by pretraining an LSTM model on resampled API and mined data \citep{xu-etal-2020-incorporating}. Subsequent work conducted an empirical study on the effectiveness of using a code generation model in an IDE plugin and find that developers largely had favorable opinions of their experience\citep{xu2021ide}. An inherent issue with the approach of \citet{xu-etal-2020-incorporating}, more fundamentally the dataset and parameters of the task, is that the intent can only contain a limited amount of information. 

\begin{figure}[ht]
    \centering
    \includegraphics[width=.9\linewidth,keepaspectratio]{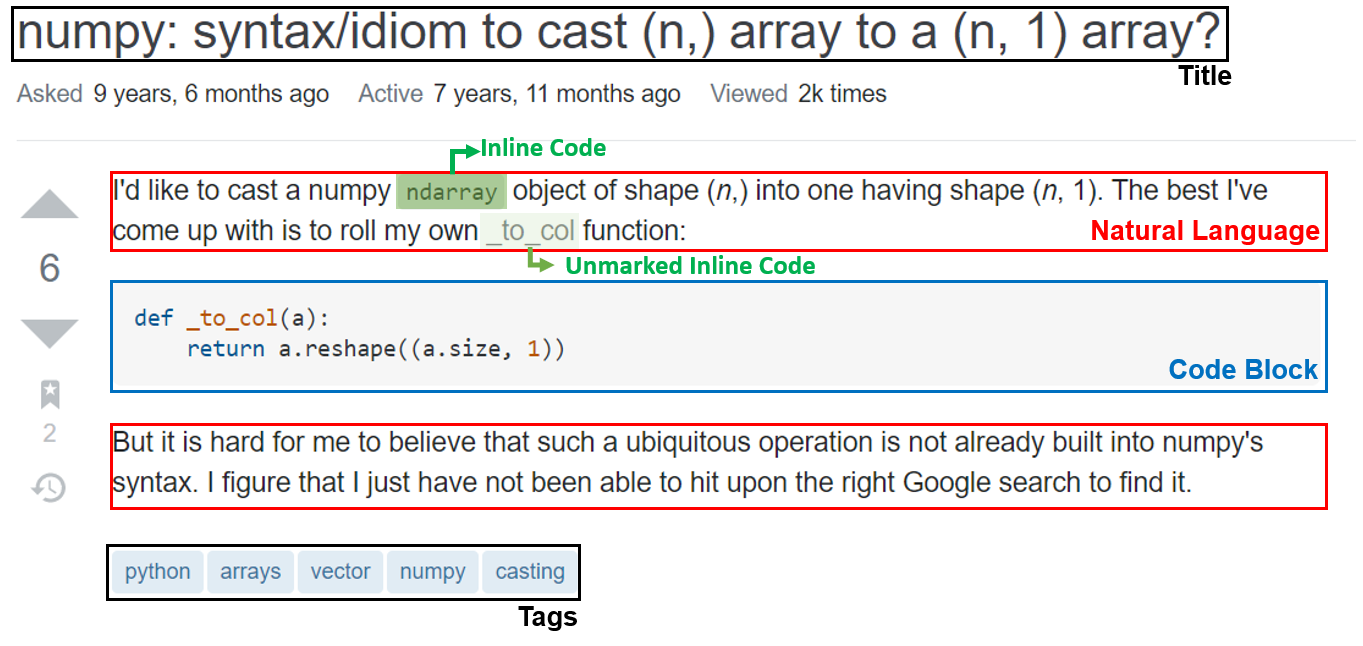}
    \caption{Example StackOverflow question with labeled elements. The corresponding rewritten intent for this question is "add a new axis to array \codesnip{a}."}
    \label{fig:examplelabeled}
\end{figure}
\indent Consider the question from \autoref{fig:examplelabeled}\textrm{ }in which a valid python snippet could be \codesnip{a[:, (np.newaxis)]}. Arriving at this answer from the intent "add a new axis to array \codesnip{a}" requires not only the disambiguation of data types for variable \codesnip{a}, but also the use of multiple distinct library-specific concepts. Further, this must be accomplished while maintaining syntactically correct code and proper order of arguments. However, neither the original title nor the rewritten intent contains the necessary information to accomplish this task. Although the previous state-of-the-art-model by \citet{xu-etal-2020-incorporating} uses abstract syntax trees (AST) to guarantee syntactically valid python code, it incorrectly generates \codesnip{a[(-1),:]=a}. One potential remedy would be to increase the amount of training data, but as discussed previously, getting high-quality annotated code generation data is especially difficult.

\indent Motivated by the limitations to the amount of information a given intent can contain and the substantial difficulty involved with gathering more labeled data, we utilize the multimodal text from the question bodies provided by the StackExchange API\footnote{https://api.stackexchange.com/}. We take advantage of the strong performances of transformer models to beat the previous state-of-the-art by 3.06 BLEU. We ensure a fair comparison by training the models from prior works with the extra data to adequately evaluate our proposed method. When all models are trained with the extra data, using BART beats the previous state of the art by 15.12 BLEU. 

\indent Our main contributions are the following:
\begin{itemize}
    \item Expanding upon the original CoNaLa dataset \citep{yin2018learning} to include the multimodal textual question bodies and thus the pertinent contextual information they contain such as inputs, outputs, and required libraries.  
    \item Demonstrating that BART does not rely on a single modality, but rather achieves its best performance on our dataset when all modalities are included. This indicates at least a shallow understanding of both natural and programming language as well as how they are related in the context of SO questions.
    \item Conducting experiments revealing that BART's struggle to generate syntacically correct code is likely a result of its tendency to be extractive rather than generative in the task of text-to-code generation.
\end{itemize}

\section{Methodology}
As detailed in \autoref{fig:approach}, our overarching approach is to: \textit{(1)} gather textual bodies from SO for both the annotated and mined examples in the CoNaLa corpus, \textit{(2)} use the concatenated intents and question bodies as inputs for a large pretrained language model, and \textit{(3)} use beam search to generate the answer code snippet.        

\subsection{StackOverflow Data}\label{subsec:sodata}
Every example $e_i \in E$ from the CoNaLa dataset \citep{yin2018learning} is comprised of an intent $x_i \in X$ that concisely summarizes what the poster wants and a snippet of Python code $y_i \in Y$ that represents an implementation of $x_i$. Crowd sourcing was used to rewrite a selection of the mined intents to reflect the snippet better and to ensure that the snippet was indeed a correct answer. As discussed, these intents are limited in the amount of information they can contain. The intent "add a new axis to array \codesnip{a}" from \autoref{fig:examplelabeled} could refer to a wide variety of different Python objects. It could range from the default \codesnip{list} to the \codesnip{Tensor} object from PyTorch\footnote{https://pytorch.org/}. The full question, or either its tags or title, is typically enough for a human to disambiguate the correct library to use. But the annotated intent lacks this crucial information as it is rather difficult to design an annotation task for SO data.\footnote{We direct readers to \citet{yin2018learning} for a full discussion of these challenges.} 

\indent We address this problem directly by using the additional data found in the SO question. In \autoref{fig:examplelabeled} there are four direct mentions of the NumPy library: two in the question body and one each in both the tags and the title. Further, there is a direct mention of the \codesnip{ndarray} data type from NumPy. It is, therefore, rather intuitive to include this additional data as input with the hope that it improves the answer generation performance. Although we did mention that both the tags and title provide salient information, the focus of this paper is only on using the noisy textual question bodies. Therefore, for every example $e_i$ the inputs now become the concatenation of $x_i$ and the body $q_{x_i} \in Q$ from the original SO question. It is important to note that $|Q|\neq |E|$ as a single question can have many examples while every question is, by definition, unique.

\subsection{Unsupervised Modality Learning}\label{subsec:unsupermodal}
 Multiple modalities are present in the textual body of a given question. These can range from embedded images to messages from administrators (or upset users) stating that the question is a duplicate of some tangentially related post that does not have an answer. While these are useful to readers, we limit our focus to three modalities: code blocks, inline code, and NL. These modalities are marked in \autoref{fig:examplelabeled} with blue, green, and red, respectively. Ideally, we would prefer to leave in the HTML tags to serve as sentinel tokens, but, looking at \autoref{fig:examplelabeled}, one immediately finds that the poster forgot to mark \codesnip{\_to\_col} as inline code. Therefore, we remove all HTML tags from the inputs, creating an unsupervised learning environment. Therefore, we propose that a transformer model will learn each of the three modalities and learn to use the relationships between each. We use BART\citep{lewis-etal-2020-bart} because its pretraining focuses on denoising textual data and, to the best of our knowledge, has minimal exposure to code examples. We used HuggingFace's~\citep{wolf-etal-2020-transformers} \codesnip{BartForConditionalGeneration} model which has a default BART encoder-decoder model with a linear layer and bias for outputs.  

\subsection{Unlabeled Data}\label{subsec:unlabeled}
We followed \citet{xu-etal-2020-incorporating} by using large amounts of the mined but not annotated data.  Unlike \citet{xu-etal-2020-incorporating}, however, we do not use this data for pretraining. Instead, we combine this data with the annotated data in our main training and validation sets. 
By adding more questions to the training set, we directly increase the probability that the model encounters a larger and more representative distribution of libraries. Intuitively, this will reduce the variances between experiments as we have reduced the dependency on the specific examples used in the training and validation sets. This variance reduction is especially useful when working with a small dataset such as CoNaLa.

\section{Experiments}
\begin{table*}[ht]
    \centering
    \begin{tabular}[c]{r|l l l l l l}\toprule
    \textbf{Split} & $|E|^*$ & $|Q|^*$ & $|E|/|Q|^\circleone$ & Intent Tokens$^\circletwo$ & Snippet Tokens$^\circletwo$ & Body Tokens$^\circlethree$\\\hline
Train & 2376 & 1708 & \numwithstd{1.39}{1.02} & \numwithstd{16.45}{7.51} & \numwithstd{17.23}{8.58} & \numwithstd{221.90}{202.65}\\
Test & 498 & 364 & \numwithstd{1.37}{0.88} & \numwithstd{15.98}{6.62} & \numwithstd{18.47}{12.90} & \numwithstd{208.04}{164.74}\\
Mined-10K & 9988$^{\thincross}$ & 7181 & \numwithstd{1.39}{0.80} & \numwithstd{11.29}{3.94} & \numwithstd{16.58}{9.27} & \numwithstd{297.53}{367.09}\\
Mined & 593837 & 40522 & \numwithstd{14.65}{7.01} & \numwithstd{11.41}{4.22} & \numwithstd{28.70}{42.81} & \numwithstd{371.24}{483.67}
    \end{tabular}
    \caption{Statistics for the CoNaLa dataset with data from the StackOverflow API. $|E|$ is \# of examples. $|Q|$ number of questions. Values are reported as $\mu \pm \sigma$ unless the column header has $^*$. $^\circleone$Mean \# of examples for a Question. $^\circletwo$Per example. $^\circlethree$ Number of tokens in the body regardless of modalitiy. $^{\thincross}$12 of the 10K questions were removed because there was an issue with them.}
    \label{tab:conaladata}
\end{table*}
% \section{Evaluation}
\begin{table*}[ht]
    \centering
    \begin{tabular}[c]{r|l l l l l l l}\toprule
    \textbf{Split} & Have Answer$^*$& Has Code & Inline$^\circleone$ & Blocks$^\circleone$& Code Tokens$^\circleone$ & NL Tokens$^\circleone$\\\hline
        Train & 87.88\% & 85.95\% & \numwithstd{1.21}{2.09} & \numwithstd{1.42}{1.26} & \numwithstd{95.54}{157.52}&\numwithstd{124.60}{92.02}\\
Test & 87.09\% & 87.91\% & \numwithstd{1.08}{1.87} & \numwithstd{1.50}{1.26} & \numwithstd{88.21}{116.01}&\numwithstd{118.52}{79.51}\\
Mined-10K & 86.16\% & 84.00\% & \numwithstd{1.30}{2.36} & \numwithstd{1.46}{1.34} & \numwithstd{133.20}{278.20}&\numwithstd{164.54}{207.08}\\
Mined & 81.92\% & 81.83\% & \numwithstd{1.50}{2.86} & \numwithstd{1.47}{1.44} & \numwithstd{172.57}{372.32}&\numwithstd{197.98}{257.71}
    \end{tabular}
    \caption{Detailed statistics for the StackOverflow questions. Mined-10K represents the top 10,000 samples selected from the Mined data based on their probability that they are a valid NL-Code pair. $^*$Percent of questions that have an accepted answer. $^\circleone$Per question body.}
    \label{tab:sodata}
\end{table*}
\subsection{Datasets}
\textbf{CoNaLa} \citep{yin2018learning}\footnote{https://conala-corpus.github.io/} is an open domain text to code generation task constructed from SO questions. It has 2,879\footnote{Actual Number is lower due to errors in the dataset preventing the usage of some examples.} annotated NL-code pairs with more than 590K mined pairs from over 40,000 unique SO questions in the dataset. 

\noindent\textbf{StackOverflow Data} For every unique question in both the annotated and mined sets, we gather additional data from the StackExchange API. As discussed in \autoref{subsec:sodata}, we only use the question body as input. Therefore the task is to generate a valid answer snippet from both the intent and the textual body. Detailed statistics for this dataset are given in \autoref{tab:conaladata} and \autoref{tab:sodata}.

\subsection{Methods}\label{subsec:methods}
We removed 238 (~10\%) examples from the training set to form the validation set. We then followed \citet{xu-etal-2020-incorporating} in taking the top mined samples based on their given probability that the NL-Code pair is valid. However, we only used 10,000 samples rather than the 100,000 \citet{xu-etal-2020-incorporating} used. From this, we remove 1000 for validation.\footnote{Some questions were deleted from StackOverflow in both the annotated and mined sets, so we could not use those.} For all tests of our model with the mined data, we combine the two training and validation sets into one. 

\indent Every experiment and test conducted in this work was conducting using Google's Colab Pro service. It afforded us the ability to use 512 input tokens with a batch size of 16. More importantly, we were able to use P100 and V100 graphics cards. Following that, we perform an ablation study using BART and the different components of our approach. Every ablation is run five separate times with different seeds and validation splits. For each test, the model with the lowest validation loss is used in the evaluation. Each test is run for ten epochs as we consistently observed overfitting after five to eight epochs.

\indent Because we introduce new data at inference, we needed to ensure we fairly compare our methods with previous work. To this end, we run the prior works with the question bodies as inputs. However, for testing \citet{xu-etal-2020-incorporating} with the question bodies, we limited the amount of mined data in pretraining to 10,000 instead of 100,000. This was done due to Google Colab's execution time limits, as it took upwards of four hours for each run of \citet{xu-etal-2020-incorporating} with only 10,000 samples. 
\subsection{Metrics}\label{subsec:metrics}
We measure the corpus level BLEU score of the generated code snippets with the same postprocessing methods and smoothing as \citet{xu-etal-2020-incorporating}. We evaluate our ablations by comparing the corpus BLEU score and unigram, bigram, and trigram precision. Finally, we calculate the percentage of test examples for which our model generated a syntactically valid Python snippet.   

\indent For the previous state-of-the-art, we also report the Oracle BLEU proposed by \citet{yin-neubig-2019-reranking}. This is calculated by choosing the candidate snippet $s_i$ with the highest sentence level BLEU score out of $n$ generated snippets. Formally, given the candidate list $C=[c_1,\ldots,c_n]$ and ground truth $y_i$, 
\begin{equation}\label{eq:oraclebleu}
    \begin{split}
        z = \underset{c_j\in C}{\textrm{argmax}} \textrm{ BLEU}(c_j,y_i)
    \end{split}
\end{equation} 

\indent Furthermore, we want to quantify how much our model relies on the body of the question or "cheats." To do this, we calculate the cheating for the generated snippet $s_i \in [s_1, \ldots, s_N]=S$ and ground truth $y_i \in [y_1,\ldots,y_N] = Y$ with respect to the input text $b_i \in [b_1,\ldots, b_N]=B$. Given the function $m(a,b)$ that calculates a textual similarity metric $m$, we define the cheating w.r.t.~$m$ as
\begin{equation}\label{eq:cheat}
    \begin{split}
    C_m(S) &=\frac{ \sum_{i\in [1;N]} (m(s_i,b_i)-m(y_i,b_i))}{N}    
    \end{split}
\end{equation}

If the model is heavily "cheating" from the input, then $m(s_i,b_i)\gg m(y_i,b_i)$, which leads to a large $C_m$. The quantity $C_m$ is, by design, similar to a standard mean squared error. The largest difference is that the difference is not squared, to facilitate distinguishing between less and more similar.

\indent For the metric function $m$, we use BLEU and ROUGE~\citep{lin-2004-rouge}. For the former, we take the bigram ($C_{BB}$) and trigram ($C_{BT}$) precision from BLEU. For ROUGE, we use bigram ROUGE (ROUGE-2/$C_{R2}$) and the longest common subsequence (ROUGE-L/$C_{RL}$). The intuition behind using these metrics is that there is a high probability that unigram precision is large. The answers to a question must address the contents of the said question, leading to shared tokens between inputs and outputs. However, the probability should massively drop when considering multiple grams. Therefore, the similarity between $n$-grams when $n>1$ should indicate the reliance on the inputs.

\subsection{Implementation}
We implemented our model with Python and HuggingFace's transformer library \citep{wolf-etal-2020-transformers}\footnote{https://github.com/huggingface/transformers}. We used a BART model with a linear layer and a separate bias for text generation. We utilized the smallest available BART model from FAIR, which was the Facebook/BART-base\footnote{https://huggingface.co/facebook/bart-base}. For training, we again rely on HuggingFace's trainer and their implementation of the learning rate scheduler. We used Adam~\citep{DBLP:journals/corr/abs-1711-05101} as our optimizer with a learning rate of \num{5e-5} and a linear learning rate scheduler. We also used a warmup ratio of 0.05. Finally, for generation, we used beam search with four beams, early stopping, and a length penalty of 0.9.

\section{Results}\label{sec:results}
\begin{table*}[h]
    \centering
    \begin{tabular}{l |l|l l}\toprule
        &\textbf{No Body}&\multicolumn{2}{c}{\textbf{With Body}}\\
        \textbf{Model} & \textbf{Corpus BLEU}&\textbf{Corpus BLEU}&\textbf{Oracle BLEU}\\
        \hline
        % Seq2Seq$^*$ & 10.58 \\
        TranX \citep{yin-neubig-2018-tranx}  & 24.30&\numwithstd{18.85}{1.26}&\numwithstd{31.21}{0.30}\\
        RR \citep{yin-neubig-2019-reranking} & 30.11&\numwithstd{19.85}{1.21}&\numwithstd{31.21}{0.30}\\
        EK \citep{xu-etal-2020-incorporating} & 30.69&\numwithstd{20.37}{1.44}&\numwithstd{33.71}{0.83}\\
        EK+RR\citep{xu-etal-2020-incorporating} & \textbf{32.26}&\numwithstd{20.54}{0.85}&\numwithstd{33.71}{0.83}\\
        \hline
        BART   & \numwithstd{26.24}{0.31}$^\circleone$&\numwithstd{34.35}{1.01}&$\geq 34.35$\\
        \textbf{BART W/ Mined} &  \numwithstd{30.55}{0.38}$^\circleone$& \textbf{\numwithstd{35.32}{0.42}}&$\geq 35.32$\\
        \bottomrule
    \end{tabular}
    \caption{Results compared to previous papers both with and without the use of the question body at inference. We do not calculate the Oracle BLEU for either of our models as our corpus BLEU already surpasses their Oracle BLEU. EK=Using External Knowledge. RR=Using Reranking.
    $^\circleone$Using only the rewritten intent, if available else normal intent, as input.
    }
    \label{tab:results}
\end{table*}
\indent We list the previous state-of-the-art BLEU scores for the CoNaLa dataset as well as the performance of our models in \autoref{tab:results}. Using the intent and question bodies achieved a BLEU score of \numwithstd{34.35}{1.01}. This was further increased to \numwithstd{35.32}{0.42} by including the mined data in the training and validation set. To better understand our model, we perform ablation tests and report their results in \hyperref[tab:ablations]{Table \ref{tab:ablations}}. When comparing our top performance with the previous top performance, regardless of the data used, our model beats the previous state of the art by 3.40 BLEU, a 10.54\% increase. Notably, our model outperforms the previous SoTA by 14.78 BLEU, a 71.96\% increase when only comparing the experiments with the question body. Furthermore, BART with the mined data and question bodies beats their Oracle BLEU by 1.61 BLEU, translating to a 4.78\% increase. However, it is important to note that \citet{xu-etal-2020-incorporating} outperforms our model by 1.71(5.30\%) when we do not use the textual body. But they still both beat the baseline TranX by 25.72\% and 7.98\%, respectively. The use of the mined data further beat the reranker by 1.46\%. 

\indent The 71.96\% increase is likely because TranX models were never intended to perform well with very noisy data, as evidenced by the 36\% dropoff in corpus BLEU when adding the body to \citet{xu-etal-2020-incorporating}. In choosing BART, we intentionally picked a transformer model designed for denoising \citep{lewis-etal-2020-bart}. Further testing is likely needed to determine if our approach is heavily dependent on the underlying transformer, but that is beyond the scope of this paper.  
\subsection{Impact of adding the Question Body}
\indent Adding the body of the question objectively improved the performance of the model. The BLEU score increased 30.92\% to 34.35 and, per \hyperref[tab:ablations]{Table \ref{tab:ablations}}, there was an increase across unigram, bigram, and trigram precision. While they all do increase, the amount is far from constant. The unigram precision only saw a 3.61\% increase, whereas bigram and trigram precision increased by 12.77\% and 22.90\%, respectively. This indicates that while the model selected slightly more correct tokens, it greatly improved its ordering of said tokens.

\indent Similar improvements, albeit smaller in value, also occurred when including the mined data without the question bodies. However, there was a sharp drop in the standard deviations for the three precision metrics. In contrast, adding the question body resulted in a steep increase in variance. This is most probably a result of the "shrinking" of the dataset that occurred when we added the bodies. In \autoref{tab:conaladata} we report that \textit{every} split of the dataset has fewer unique questions than it does examples. Also reported is that the number of tokens in the body is, on average, significantly greater than that of the intents. The effective dataset size is now much smaller, while the number of unique answer snippets stayed the same. The result is that the model now performs better on the difficult test set, at the cost of being more reliant on the training and validation split. Using both the bodies and mined data does mitigate this "shrinking" effect, as shown by the lower standard deviations than those when only using the body.

\begin{table*}[h]
    \centering
    \begin{tabular}{l|l|lll|ll}
        \toprule
        \textbf{Input}&  \textbf{BLEU} &  \textbf{Unigram}$^*$ &  \textbf{Bigram}$^*$ & \textbf{Trigram}$^*$ & \textbf{Valid}$^\circleone$\\
        \hline
        Baseline  & \incorrettext{\numwithstd{26.24}{0.31}} & \numwithstd{67.53}{0.46} & \incorrettext{\numwithstd{44.10}{0.60}} & \incorrettext{\numwithstd{29.80}{0.69}} & \numwithstd{84.08}{1.27}\\
        \tableind +Mined & \numwithstd{30.55}{0.38} & \numwithstd{67.81}{0.23} & \numwithstd{45.55}{0.27} & \numwithstd{31.69}{0.37} & \textbf{\numwithstd{93.08}{1.28}}\\
        \hline
        Body & \numwithstd{34.35}{1.01} & \numwithstd{69.97}{0.89} & \textbf{\numwithstd{49.74}{0.99}} & \textbf{\numwithstd{36.62}{0.97}} & \numwithstd{81.44}{2.25}\\
        \tableind -NL & \numwithstd{34.06}{0.48} & \numwithstd{68.29}{0.48} & \numwithstd{47.91}{0.45} & \numwithstd{35.33}{0.40} & \numwithstd{81.92}{0.75}\\
        \tableind -Code & \numwithstd{27.67}{0.40} & \numwithstd{68.29}{0.53} & \numwithstd{44.93}{0.57} & \numwithstd{30.12}{0.69} & \numwithstd{84.92}{1.00}\\
      \tableind -Blocks & \numwithstd{29.53}{0.47} & \numwithstd{68.14}{0.26} & \numwithstd{45.69}{0.10} & \numwithstd{31.36}{0.15} & \incorrettext{\numwithstd{80.84}{1.37}}\\
        \tableind -Inline & \numwithstd{33.57}{0.94} & \textbf{\numwithstd{70.50}{0.27}} & \numwithstd{49.56}{0.40} & \numwithstd{36.54}{0.46} & \numwithstd{82.16}{1.53}\\
        \hline
        \textbf{Body+Mined} & \textbf{\numwithstd{35.32}{0.42}} & \numwithstd{67.62}{0.76} & \numwithstd{47.69}{0.82} & \numwithstd{35.00}{0.87} & \numwithstd{89.32}{1.49}\\
        \tableind -NL & \numwithstd{34.53}{0.88} & \incorrettext{\numwithstd{66.24}{0.90}} & \numwithstd{46.11}{1.15} & \numwithstd{33.54}{1.02} & \numwithstd{90.08}{0.48}\\
        \tableind -Code & \numwithstd{31.39}{0.75} & \numwithstd{67.00}{0.75} & \numwithstd{45.65}{0.97} & \numwithstd{31.60}{0.88} & \numwithstd{92.00}{1.31}\\
    	\tableind -Blocks & \numwithstd{32.14}{0.14} & \numwithstd{66.96}{1.03} & \numwithstd{45.32}{0.97} & \numwithstd{31.49}{0.74} & \numwithstd{89.24}{1.30}\\
        \tableind -Inline & \numwithstd{35.06}{0.49} & \numwithstd{67.04}{1.54} & \numwithstd{46.99}{1.29} & \numwithstd{34.31}{1.04} & \numwithstd{89.20}{0.42}\\
        \bottomrule
    \end{tabular}
    
    \caption{Ablation Experiments all with BART ran on 5 different random initializations. All tests have rewritten intent as input in addition to the input described in the \textbf{Input} column. The \textbf{bolded ablation} indicates our best performance while \incorrettext{red text} represents the worst performance. $^*$Precisions. $^\circleone$Percent of generated snippets that are valid python.}
    
    \label{tab:ablations}
\end{table*}
\subsection{Is BART Reliant on a Single Modality} \label{subsec:bartreliant}
\begin{table}[]
    \centering
    \begin{tabular}{l|rr}\toprule
         &  Body& Body+Mined \\
         \hline
         -NL & -0.29&  -0.80\\
         -Code &  -6.68&-3.93 \\
         -Blocks&  -4.83 &-3.18\\
         -Inline & -0.79&-0.26\\
         \bottomrule
    \end{tabular}
    \caption{Change in BLEU score for each ablation versus their respective baseline.}
    \label{tab:ablationdrops}
\end{table}
As discussed in \autoref{subsec:unsupermodal}, we focus on three modalities in the textual bodies: code blocks, inline code, and natural language. We put forth the idea that a large pretrained language model such as BART learns each modality in an unsupervised manner. We designed four distinct ablations to test if this was the case. Each was run both with and without the mined data totaling eight ablations. We report the full BLEU scores from these in \autoref{tab:ablations}. Further, we calculate the performance with respect to baselines in \autoref{tab:ablationdrops}. Notably, there was no modality whose removal resulted in a BLEU score worse than when the question body was not used in the input. There was also not a modality whose removal improved performance. From our ablations, it is clear that the most important modality in the question bodies is the code regardless of if it is inline or in a block. But, using only code is still 2.25\% worse than when all three modalities are included with mined. This indicates that the NL surrounding acts not only as additional context, but likely further both direct and indirect indicators of salient code for the model.

\subsection{Removing Code Improves Syntax}\label{subsec:quality}

In \autoref{tab:ablations} we report the percent of generated snippets that are syntactically valid—adding only the mined data results in a 9\% increase. When using the question bodies, the addition of the mined data also increases the percent of valid snippets generated by 7.88\%. While it is an improvement, it is still a 3.76\% drop from when the body was excluded. Further, removing the code from the bodies resulted in the highest percentages of 92.00\% and 84.92\% with and without the mined data. We then performed a finer analysis using a single seed and the same training and validation data across all ablations and reported the results in \autoref{app:errorcats}. Across all ablations, the majority of errors are caused by mismatches of parentheses. In reality, a large percentage of general syntax errors are likely caused by this. However, syntax errors prevent the extraction of the AST for further investigation of these errors.

\indent We also report in \autoref{tab:pctprint} the percentage of valid snippets generated when the \codesnip{print} function is present. One of the more commonly occurring incompatibilities between Python 2 and 3 is that \codesnip{print} now requires parentheses. Considering that the questions in the CoNaLa dataset are from March 2017 or earlier~\citep{yin2018learning} and that support for Python 2.x only ended in January 2020\footnote{https://www.python.org/doc/sunset-python-2/}, we hypothesize that these deprecated calls are a large cause of the errors. When both the body and snippet have \codesnip{print}, the inclusion of the question body led to the percent of valid snippets dropping by 21.06 with and 21.05 without the mined data with respect to their baselines. While there are only 19 such questions in the test set, this is a significant drop. The likely cause is that the autoregressive decoder of BART struggles to remember to close the parentheses when wrapping the snippet with a \codesnip{print} statement. One solution would be to run the \codesnip{2to3}\footnote{https://docs.python.org/3/library/2to3.html} translator on all of the code. However, the possibilities for code blocks to contain code and other modalities such as error messages and console executions present significant hurdles as \codesnip{2to3} does not support these. Therefore we leave that to future work.

\subsection{Cheating}
In \autoref{subsec:metrics} we define the "cheating" equation to measure if the generated snippet is more similar to the question body than the ground truth is. The ideal model would maximize the BLEU score while minimizing the $|C_{m}|$. We run multiple ablations on a single seed and calculate the "cheating" as defined by \autoref{eq:cheat} and present these results in \autoref{tab:cheatres}.

\begin{table}[h]
    \centering
    \begin{tabular}{l|rrrr}
        \toprule
         & $C_{BB}$&$C_{BT}$&$C_{R2}$&$C_{RL}$ \\
        \hline
        Baseline & -6.82& -6.91 & -1.62 & -1.98\\
        \tableind +Mined &-6.03&-6.03&-1.41&-0.29\\
        \hline
        Body  & 11.65 & 13.37 & 1.75 & 1.18 \\
        \tableind -Code & -4.18 & -5.05 & -1.06 & -1.40\\
        \tableind -NL & 10.55 & 12.27& 1.24 & 0.47 \\
        \tableind -Blocks &-3.32 & -3.48&-0.89&-1.09\\
        \tableind -Inline & 9.19& 10.39& 0.90 & 0.44\\
        \hline
        Body+Mined &10.19 & 12.55 &1.39 & 0.48\\
        \tableind -Code &-4.37 & -5.05&-1.08&-1.47\\
        \tableind -NL & 9.40&11.32&1.19 & 0.17\\
        \tableind -Blocks &-3.48 & -4.16&-0.84&-1.19\\
        \tableind -Inline & 7.93 & 9.73 & 1.11 & 0.25\\
    \end{tabular}
    \caption{Cheating Measurements calculated by \autoref{eq:cheat} using a single run but same seed and environment. $C_{BB}\textrm{ and }C_{BT}$ are the cheating w.r.t.~BLEU Bigram and Trigram Precision. $C_{R2}\textrm{ and }C_{RL}$ are the cheating w.r.t.~ROUGE-2 and ROUGE-L.}
    \label{tab:cheatres}
\end{table}

\indent Suffice to say that \textit{serious} violations of academic integrity have occurred. As expected, the baseline is less similar to the question bodies than the ground truth is. When the body was used as input, $C_{BT}$ increased by 20.28 points, while $C_{RL}$ rose by 3.16 points, representing a 293.49\% and 159.60\% increase over their respective baselines. Including the mined data resulted in increases of 18.59 (308.13\%) and 0.77(265.52\%) when compared to using only the intents. Both indicate that the model's generated output has significantly more shared multigram subsequences with the question body than the ground truth does. In the ablations where code was removed from the body, $C_{BT}$ increased by only 0.98 and 1.86 with and without the mined data. This represents a percent of increase of only 16.25\% and 26.92\% over their respective baselines. However, in the case where all NL was removed, $C_{BT}$ increased by 17.35(287.73\%) and 19.18(277.57\%) points with respect to their baselines. The fact that these increases are lower than that when all modalities are included provides further evidence that BART is an unsupervised multimodal learner and understands the relationships between each modality. The NL likely provides both explicit and implicit hints about the importance of certain code spans.

\subsection{Examples}
\begin{table}[h]
    \centering
    \begin{tabular}{P{7.5cm}}
    \toprule
    \small\textbf{Intent:} multiply a matrix `p` with a 3d tensor `t` in scipy\normalsize\\\hline
     \checkding\textrm{ }\codetable{scipy.tensordot(P, T, axes=[1, 1]).swapaxes(0, 1)}\\
     \circleone\textrm{ }\codetable{\incorrettext{np.einsum(`...j,...j->...`,} \incorretplace{P, T}\incorrettext{)}}\\
     \circletwo\textrm{ }\codetable{\incorrettext{np.einsum('ij->ij->ik->j->ik', p)}}\\
     \circleone\textrm{ }\codetable{\incorretplace{P}\incorrettext{.dot(}\incorretplace{T}\incorrettext{).transpose(1, 0, 2)}}\\
     \hline
     \hline
     \small\textbf{Intent:} concatenate items of list `l` with a space ' '\normalsize\\\hline
     \checkding\textrm{ }\codetable{print(' '.join(map(str, l)))}\\
     \circleone\textrm{ }\codetable{\incorrettext{list}(map(\incorrettext{tuple},\incorrettext{[]}))}\\
     \circletwo\textrm{ }\codetable{\incorrettext{[item for item in L if '' in item]}}\\
     \circlethree\textrm{ }\codetable{print(' '.join(\incorretplace{str}\incorrettext{(x) for x in L}))}\\
     \hline
     \hline
     \small\textbf{Intent:} concatenate elements of list `b` by a colon ":"\normalsize\\\hline
     \checkding\textrm{ }\codetable{""":""".join(str(x) for x in b)}\\
     \circleone\textrm{ }\codetable{\incorrettext{[` `}.join(\incorretplace{x}) for x in b\incorrettext{]}}\\
     \circletwo\textrm{ }\codetable{\incorrettext{b = [int(i)} for \incorrettext{i} in b]}\\
     \circlethree\textrm{ }\codetable{\incorrettext{print('}\incorretplace{:}\incorrettext{'}.join(map(str, b))}\\     \bottomrule
    \end{tabular}
    \caption{Example intents and generated snippets. Screenshots of the questions are located in \autoref{subsec:exampleimage} and each intent links to the question. \incorrettext{Red text} indicates that it is incorrect while \incorretplace{blue text} marks correct tokens in the wrong place. \checkding ground truth. \circleone EK+RR no body~\citep{xu-etal-2020-incorporating}. \circletwo Mined. \circlethree Body+Mined.}
    \label{tab:examples}
\end{table}

We select three examples that demonstrate the benefits of our approach while also highlighting the issues in both the use of the question body and SO corpora in general and report them in \autoref{tab:examples}. In the first example, we can see that both \circleone\textrm{ }and \circletwo\textrm{ }have learned how to use einsums, but neither is correct. \circlethree\textrm{ }in this case produces an answer that returns the correct value. It is highly probable that BART understood from the poster's explicit mention that \codesnip{P.dot(T).transpose(1, 0, 2)} gives the desired result and thus extracts it. However, this example has two critical issues: the poster's intent is to find a "cleaner" way to multiply a matrix with a tensor, and \codesnip{scipy.tensordot} is deprecated. The latter is to be expected, considering the answer is from 2010. But it does indicate that a better evaluation based on inputs and outputs is likely needed.

\indent The next two examples are quite similar but are from two separate questions. \circleone\textrm{ }likely mistakes the core intent to be type conversion due to the inclusion of the words "items" and "with." \circletwo\textrm{ }also suffers from the inclusion of these tokens but believes the problem involves filtering. In the final example, \circleone\textrm{ }recognizes that it must convert the items in \codesnip{b} to \codesnip{str}, but does not return a joined string. \circletwo\textrm{ }recognizes that, again, the answer involves type conversion but predicts the incorrect type.

\indent Similar to the first example, \circlethree\textrm{ }produces answers for both the second and third examples that functionally return the correct results. However, running \circlethree's solution for the third example would result in a syntax error due to the missing "\codesnip{)}." On further inspection of the question bodies, it becomes apparent that the probable reason why one snippet is syntactically valid while the other is not is the presence of a Python 2 \codesnip{print}. The model recognizes that a suitable answer can be found in the question but must be converted to python 3. As discussed in \autoref{subsec:quality}, these print statements are prone to cause syntactical issues.   

\section{Conclusion}
We expand the CoNaLa dataset by adding the textual question bodies from the StackExchange API and achieve state-of-the-art performance with a simple BART model. Further, we demonstrate that, for this task, BART performs best when code blocks, inline code, and NL are all present. We then examine the impact of the question body on syntax errors and BART's cheating through multimodal understanding. Finally, we examine examples that highlight the issues with both StackOverflow data and code evaluation in general. Future work should focus on extracting desired inputs and outputs for a given intent. Further, additional efforts put into creating corpora of executable code are likely to improve not only generation but evaluation. Both will also protect datasets from deprecated functions and abandoned libraries.    

\bibliography{anthology,bibliography}
\bibliographystyle{acl_natbib}

\appendix

\section{Error Categories}\label{app:errorcats}
% This is an appendix.
\begin{table*}[h!]
    \centering
 \begin{tabular}{l|r|rrr}
        \toprule
          &Error Count &  General Invalid Syntax &  Paranthesis Matching &  Other Matching \\
        \hline
        Baseline  &     61 &                   39.34 &                 45.90 &           14.75 \\
        \tableind +Mined &     38 &                   47.37 &                 39.47 &           13.16 \\
        \hline
        Body   &    104 &                   39.42 &                 46.15 &           14.42 \\
        \tableind -Code  &     82 &                   30.49 &                 60.98 &            8.54 \\
        \tableind -NL  &     75 &                   25.33 &                 53.33 &           21.33 \\
        \tableind -Blocks &     94 &                   36.17 &                 53.19 &           10.64 \\
        \tableind -Inline  &     85 &                   28.24 &                 61.18 &           10.59 \\
        \hline
        Body+Mined  &     59 &                   38.98 &                 49.15 &           11.86 \\
        \tableind -Code  &     52 &                   26.92 &                 67.31 &            5.77 \\
        \tableind -NL &     48 &                   33.33 &                 47.92 &           18.75 \\
        \tableind -Blocks &     51 &                   23.53 &                 66.67 &            9.80 \\
        \tableind -Inline &     51 &                   41.18 &                 54.90 &            3.92 \\
        \bottomrule
    \end{tabular}
    \caption{Percentages of syntax errors for ablations in a single run. }
    \label{tab:pctsyntax}
\end{table*}

\begin{table*}[h!]
    \centering
     \begin{tabular}{l|rrrr}
        \toprule
          &  No Print &  Has Print in Snippet &  Has Print in Body &  Has Print in Both \\
        \hline
        Baseline &     88.28 &                 84.62 &              86.59 &              84.21 \\
        \tableind +Mined  &     92.97 &                100.00 &              91.46 &              78.95 \\
        \hline
        Body  &     78.91 &                 84.62 &              82.93 &              63.16 \\
        \tableind -Code &     84.38 &                 92.31 &              80.49 &              73.68 \\
        \tableind -NL &     84.38 &                 84.62 &              89.02 &              78.95 \\
        \tableind -Blocks &     82.29 &                 92.31 &              76.83 &              68.42 \\
        \tableind -Inline &     83.07 &                 76.92 &              86.59 &              68.42 \\
        \hline
        Body+Mined &     90.89 &                 92.31 &              81.71 &              57.89 \\
        \tableind -Code &     91.67 &                 69.23 &              84.15 &              84.21 \\
        \tableind -NL &     89.84 &                100.00 &              91.46 &              89.47 \\
        \tableind -Blocks &     90.36 &                 92.31 &              92.68 &              63.16 \\
        \tableind -Inline &     90.89 &                 92.31 &              87.80 &              73.68 \\
\bottomrule
    \end{tabular}
    \caption{Percentage of valid snippets based on the presence of \codesnip{print}.}
    \label{tab:pctprint}
\end{table*}

% \vspace{\baselineskip}
\newpage
\section{Full Questions for Examples}\label{subsec:exampleimage}
\begin{figure*}[hb!]
    
    \centering
     \begin{subfigure}[h]{.6\linewidth}
     
        \includegraphics[width=\textwidth,keepaspectratio]{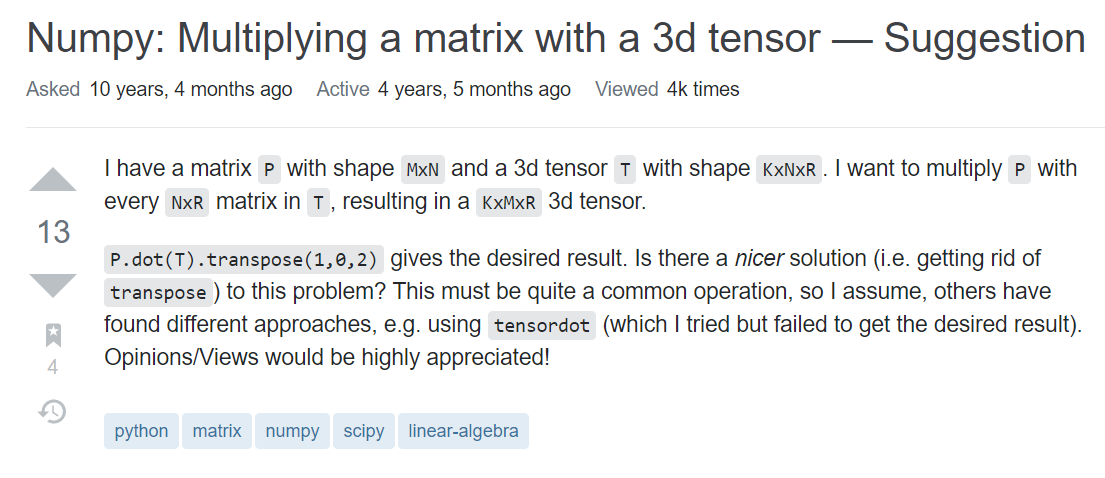}
        \caption{Full Stack Overflow Question for Example 1 in \autoref{tab:examples}. Question can be found https://stackoverflow.com/questions/4490961/numpy-multiplying-a-matrix-with-a-3d-tensor-suggestion.}
        \label{fig:example1}
     \end{subfigure}
    \vfill
    \vfill
    \begin{subfigure}[h]{.6\linewidth}
         \centering
         \includegraphics[width=\textwidth,keepaspectratio]{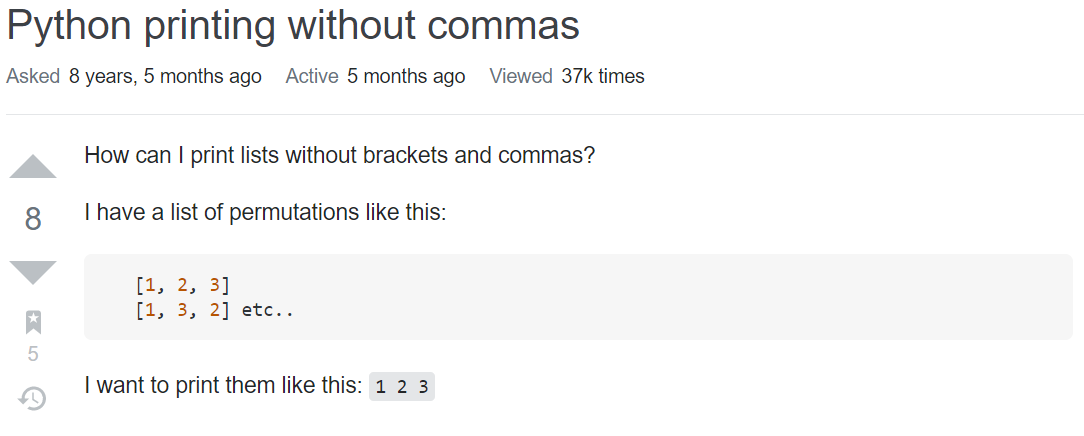}
         \caption{Full Stack Overflow Question for Example 2 in \autoref{tab:examples}. Question can be found https://stackoverflow.com/questions/13550423/python-printing-without-commas.}
         \label{fig:example2}
     \end{subfigure}
     \vfill
     
    \begin{subfigure}[h]{.6\linewidth}
         \centering
         \includegraphics[width=\textwidth,keepaspectratio]{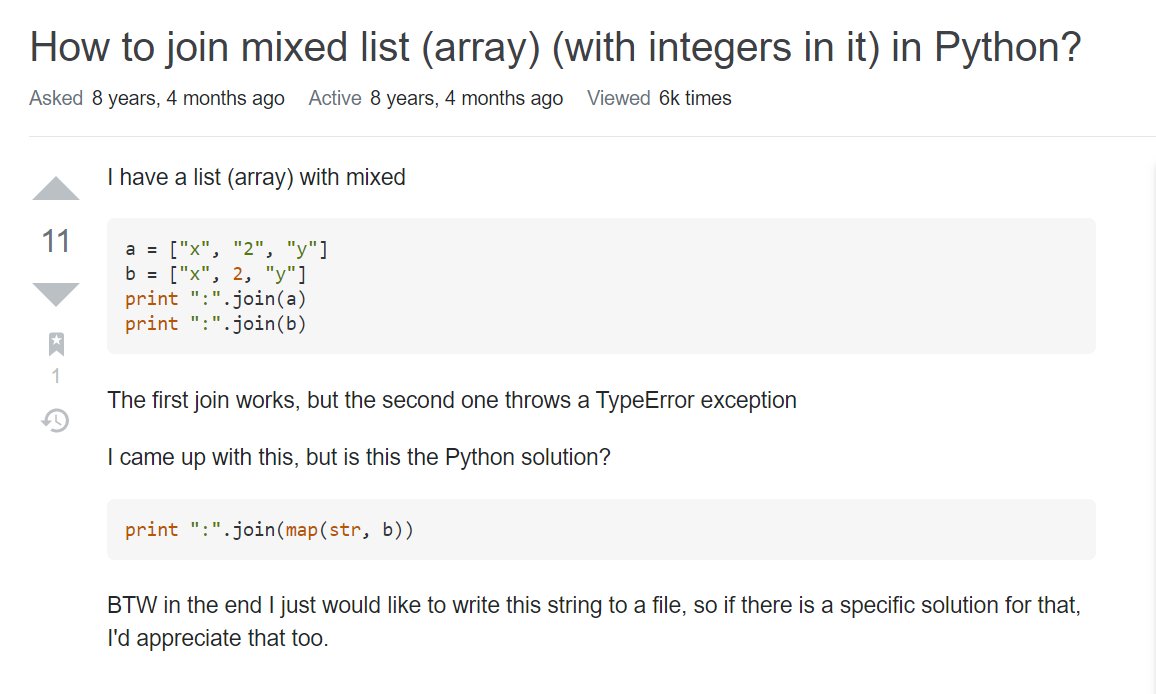}
         \caption{Full Stack Overflow Question for Example 3 in \autoref{tab:examples}. Question can be found https://stackoverflow.com/questions/13954222/how-to-join-mixed-list-array-with-integers-in-it-in-python.}
         \label{fig:example3}
     \end{subfigure}
\end{figure*}
\end{document}